\pdfoutput=1

\documentclass[11pt]{article}

\usepackage{acl}

\usepackage{xcolor}
\usepackage[most]{tcolorbox}
\usepackage{times}
\usepackage{makecell}
\usepackage{latexsym}
\usepackage{booktabs}
\usepackage{amsmath} 
\usepackage{multirow}
\usepackage{algorithm}
\usepackage{algpseudocode}
\usepackage[T1]{fontenc}

\usepackage[utf8]{inputenc}

\usepackage{microtype}

\title{E-Bench: Towards Evaluating the Ease-of-Use of Large Language Models}

\author{
Zhenyu Zhang$^\dag$\thanks{~~~Equal contribution}~, Bingguang Hao$^{\ddag*}$, Jinpeng Li$^\ddag$, Zekai Zhang$^\ddag$, Dongyan Zhao$^\ddag$ \\
$^\dag$Baidu Inc. \\
$^\ddag$Wangxuan Institute of Computer Technology, Peking University \\
\texttt{zhangzhenyu07@baidu.com, lijinpeng@stu.pku.edu.cn} \\
}

\begin{document}
\maketitle
\begin{abstract}
Most large language models (LLMs) are sensitive to prompts, and another synonymous expression or a typo may lead to unexpected results for the model. Composing an optimal prompt for a specific demand lacks theoretical support and relies entirely on human experimentation, which poses a considerable obstacle to popularizing generative artificial intelligence.
However, there is no systematic analysis of the stability of LLMs in resisting prompt perturbations in real-world scenarios.
In this work, we propose to evaluate the ease-of-use of LLMs and construct E-Bench, simulating the actual situation of human use from synonymous perturbation (including paraphrasing, simplification, and colloquialism) and typographical perturbation (such as typing). On this basis, we also discuss the combination of these two types of perturbation and analyze the main reasons for performance degradation.
Experimental results indicate that with the increase of model size, although the ease-of-use are significantly improved, there is still a long way to go to build a sufficiently user-friendly model.
\end{abstract}

\section{Introduction}

Large language models (LLMs) have swept across the entire natural language processing (NLP) fields, revolutionizing many domains and attracting unprecedented attention.
Benefiting from the powerful instruction following and language generation capabilities, the application scenarios of modern LLMs are gradually expanding and being used as productivity tools.
By exploring natural language exquisite prompts, LLMs could handle various demands, such as question answering, data construction, and intelligent agent~\cite{jakesch2023co,wu2023exploring,xu2023wizardlm}.

\begin{figure}
    \centering
    \includegraphics[width=0.95\linewidth]{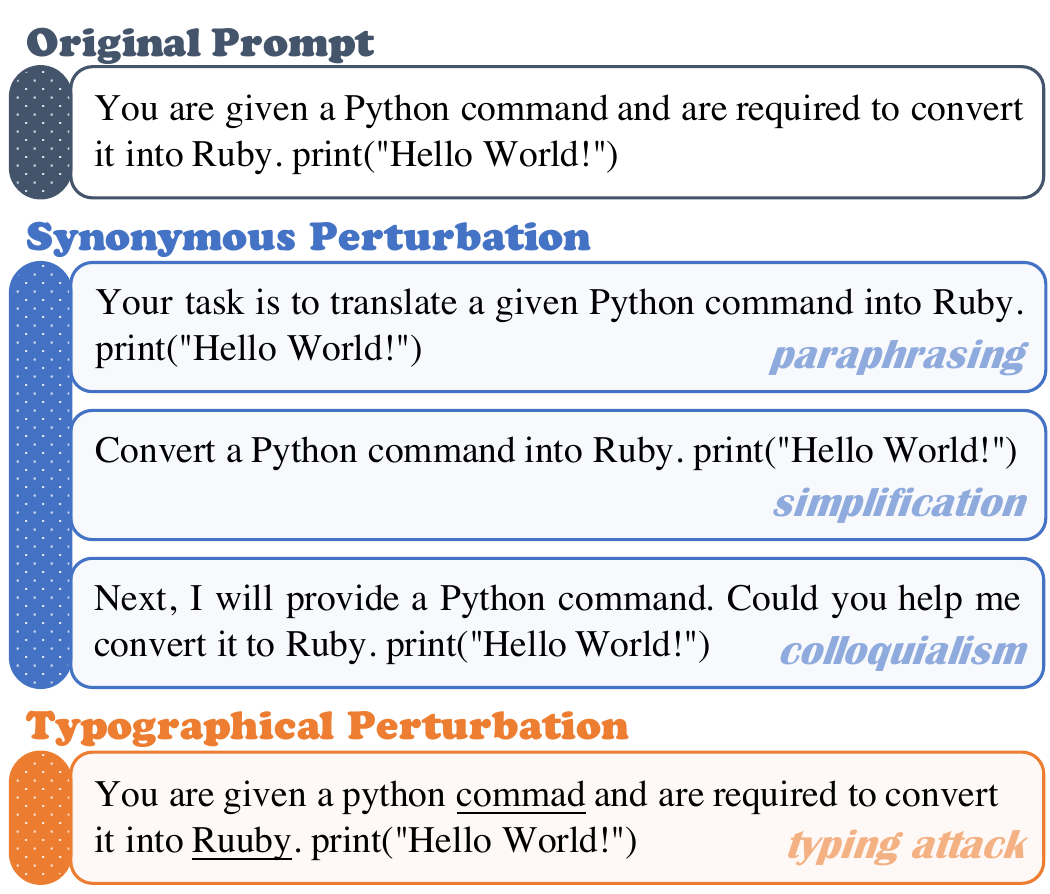}
    \caption{The prompt perturbations in E-Bench, which simulate the actual situations of humans using LLMs.}
    \label{fig:intro}
\end{figure}

Yet, LLMs are prompt-driven, the performance of downstream tasks significantly depends on the quality of the prompt used to steer the model.
Inappropriate prompts are insufficient to meet target tasks, and most effective prompts are handmade by humans~\cite{zhou2022large,kaddour2023challenges,sclar2024quantifying}.
In this process, there is no reliable theoretical basis or prior indication of what kind of prompt is optimal for a specific task. Users write a prompt, verify its validity, and refine the prompt iteratively, like opening a ``surprise box''. 
It even gives birth to a new field, prompt engineering, which attempts to explore how to write prompts to improve the efficiency of direct interaction between humans and deep generative models~\cite{cao2023beautifulprompt,chen2023unleashing,gu2023systematic}.

The sensitivity of LLMs to prompt is a huge obstacle in their usage. For example, no one knows what kind of performance a good prompt on ChatGPT will have on GPT-4.
In contrast, human conversational abilities are general, flexible, and robust, different expressions with the same semantics will not confuse different people~\cite{peng2021raddle}.
Recently, there have been some works on evaluating the robustness of LLMs~\cite{wang2023robustness,zhu2023promptbench,mousavi2024llms}. However, they mainly focus on the perspective of out-of-distribution problems or adversarial prompts of NLP tasks, rather than prompt perturbations in the most commonly used conversational scenarios, which is precisely the key to pushing LLMs away from tedious prompt engineering.
In other words, they still have not answered well whether a model is convenient enough for human use.
A robust LLM should be able to consistently provide accurate and relevant responses across a series of synonymous prompts, which helps improve the usability.

In this paper, we fuel research in this direction by introducing E-Bench, a comprehensive benchmark designed for understanding the impact of prompt perturbation and evaluating the ease-of-use of LLMs.
Here, an ideal model is expected to handle synonymous and even typographical prompts like humans swimmingly.
Starting from existing evaluation set AlpacaEval~\cite{alpacaeval}, we divide it into four parts according to data characteristics and perturb the prompts through paraphrasing, simplification, colloquialism, and typing attacks.
Figure~\ref{fig:intro} provides an illustration for each perturbation.
Specifically, we first perturb each prompt with a series of automatic tools, then manually review to ensure that the prompts before and after perturbation have similar semantics and do not affect human understanding.
Performance drop is used as evaluation metric, where the closer the performance on perturbed prompts is to original performance, the higher ease-of-use of the model.


With E-Bench, we conduct experiments to evaluate the ease-of-use of 6 representative LLMs, including Llama2-chat models~\cite{touvron2023llama2}, Vicunas~\cite{vicuna2023}, and GPTs~\cite{achiam2023gpt}.
The results demonstrate that all models experience varying performance degradation after prompt perturbation, and larger models perform better under synonymous perturbation, while there is no clear scaling law for model size and performance degradation under typing attack. Overall, improving the ease-of-use of LLMs is an urgent research topic.
Furthermore, additional analysis reveals the impact of training data on the specific aspect of ease-of-use. We hope that E-Bench could provide a stepping stone for the popularization of general artificial intelligence (GAI).

\begin{table}[!ht]
\tiny
\centering
\resizebox{\linewidth}{!}{
\begin{tabular}{llccc}
\toprule
\multirow{3}{*}{Dataset} & & \multirow{3}{*}{\#Ins} & \multicolumn{2}{c}{Avg. Len.} \\
\cmidrule(r){4-5} & & & Orig. & Pert. \\
\midrule
\multicolumn{2}{c}{AlpacaEval~\cite{alpacaeval}} & 805 & 28.83 & - \\
\midrule
\multirow{5}{*}{E-Bench} & Paraphrasing & 200 & 25.22 & 24.56 \\
& Simplification & 200 & 47.21 & 32.28 \\
& Colloquialism  & 200 & 15.66 & 20.26 \\
& Typing Attack  & 205 & 27.28 & 27.30 \\
& Typing Attack$^*$  & 805 & 26.10 & 26.73 \\
\bottomrule
\end{tabular}
}
\caption{Statistics of E-Bench, in which typing attack has its own evaluation set and could also be combined with other perturbations (marked by $*$).}
\label{tab:stat}
\end{table}

\begin{table*}[ht]
\tiny
\centering
\resizebox{\textwidth}{!}{
\begin{tabular}{lcccccccccccc}
\toprule
& \multicolumn{3}{c}{\textit{Paraphrasing}} & \multicolumn{3}{c}{\textit{Simplification}} &  \multicolumn{3}{c}{\textit{Colloquialism}} & \multicolumn{3}{c}{\textit{Overall}}  \\
 \cmidrule(r){2-4} \cmidrule(r){5-7} \cmidrule(r){8-10} \cmidrule(r){11-13} Models & Orig. & Pert. & $\Delta$ & Orig. & Pert. & $\Delta$ & Orig. & Pert. & $\Delta$ & Orig. & Pert. & $\Delta$ \\ 
\midrule
Llama 2-chat {(7b)}  & 71.00 & 62.00 & 9.00 & 69.00 & 62.31 & 6.69 & 72.50 & 56.50 & 16.00 & 70.83 & 60.27 & 10.56 \\
Llama 2-chat {(13b)} & 81.00 & 78.00 & 3.00 & 76.38 & 74.37 & 2.01 & 79.00 & 69.00 & 10.00 & 78.79 & 73.79 & 5.00 \\
\midrule
Vicuna-v1.5 {(7b)}  & 77.50 & 70.00 & 7.50 & 73.62 & 67.50 & 6.12  & 73.25 & 72.11 & 1.14 & 74.79 & 69.87 & 4.92 \\
Vicuna-v1.5 {(13b)} & 80.50 & 74.50 & 6.00 & 79.75 & 81.25 & -1.50 & 82.25 & 79.65 & 2.60 & 80.83 & 78.47 & 2.37 \\
\midrule 
GPT-3.5 {(24.01)} & 85.93 & 84.00 & 1.93 & 92.21 & 84.50 & 7.71 & 89.75 & 83.25 & 6.50 & 89.27 & 83.92 & 5.38 \\
GPT-4 {(24.01)} & 95.48 & 94.25 & 1.23 & 94.00 & 90.50 & 3.50 & 92.25 & 89.75 & 2.50 & 93.91 & 91.50 & 2.41 \\
\midrule 
Overall & 81.90 & 77.13 & 4.78 & 80.83 & 76.74 & 4.09 & 81.50 & 75.04 & 6.46 & 81.41 & 76.30 & 5.11 \\
\bottomrule
\toprule
& \multicolumn{2}{c}{\textit{Paraphrasing}} & \multicolumn{2}{c}{\textit{Simplification}} &  \multicolumn{2}{c}{\textit{Colloquialism}} & \multicolumn{3}{c}{\textit{Typing Attack}} & \multicolumn{3}{c}{\textit{Overall}}  \\
\cmidrule(r){2-3} \cmidrule(r){4-5} \cmidrule(r){6-7} \cmidrule(r){8-10} \cmidrule(r){11-13} Models & +Att. & $\Delta$ & +Att. & $\Delta$ & +Att. & $\Delta$ & Orig. & +Att. & $\Delta$ & Orig. & +Att. & $\Delta$ \\ 
\midrule
Llama 2-chat {(7b)}  & 33.17 & 28.83 & 30.50 & 31.81 & 33.84 & 22.66 & 73.28 & 59.50 & 13.78 & 63.52 & 39.25 & 24.27 \\
Llama 2-chat {(13b)} & 39.50 & 38.50 & 42.50 & 31.87 & 38.50 & 30.50 & 75.85 & 60.98 & 14.87 & 74.31 & 45.37 & 28.94 \\
\midrule
Vicuna-v1.5 {(7b)}  & 42.50 & 27.50 & 41.50 & 26.00 & 49.50 & 22.61 & 75.36 & 61.71 & 13.65 & 71.24 & 48.80 & 22.44 \\
Vicuna-v1.5 {(13b)} & 44.50 & 30.00 & 45.00 & 36.25 & 50.25 & 29.40 & 84.30 & 71.50 & 12.80 & 79.93 & 52.81 & 27.11 \\
\midrule 
GPT-3.5 {(24.01)} & 50.00 & 34.00 & 51.50 & 33.00 & 58.75 & 24.50 & 93.41 & 82.61 & 10.80 & 86.29 & 60.72 & 25.57 \\
GPT-4 {(24.01)}  & 58.00 & 36.25 & 63.50 & 27.00 & 68.00 & 21.75 & 93.96 & 88.65 & 5.31 & 92.12 & 69.54 & 22.58 \\
\midrule 
Overall & 44.61 & 32.51 & 45.75 & 30.99 & 49.81 & 25.24 & 82.69 & 70.83 & 11.87 & 77.90 & 52.75 & 25.15 \\
\bottomrule
\end{tabular}
}
\caption{Main results on E-Bench, where `Orig.', `Pert.', and '+Att.' refer the performance of on original, perturbed, and attacked prompts, respectively, `$\Delta$' indicates the preference drop. See Appendix~\ref{appd:annotation} for statistical analysis.}
\label{tab:main}
\end{table*}

\section{E-Bench}

Inspired by actual scenarios, we first consider different expressions of the same semantics, including \textit{paraphrasing}, \textit{simplification}, and \textit{colloquialism}.
In practical terms: 1) paraphrasing is the simplest form of expression conversion and will not significantly impact the length of prompt. 2) simplification is a quality in which users express their intents using fewer words to be concise. 3) colloquialism converts written style prompts into spoken forms, usually with modal particles and increased length.
Furthermore, we simulate the typo issues during prompt input and carry out char/word-based attacks on the original input or after each synonymous perturbation above.

\subsection{Construction}

We reconstruct AlpacaEval~\cite{alpacaeval}, an automatic evaluator for instruction-following models with 805 examples, to perform the ease-of-use evaluation.
The dataset is divided into four parts. First, 205 instances are randomly selected for typing attack. Then, the remaining 600 instances are arranged according to prompt length and divided into three parts: the shortest part is used for colloquialism, the longest part is for simplification, and the part with moderate length is for paraphrasing.


\noindent \textbf{Paraphrasing.}
It exists widely because different users have different language habits. 
Here, we focus on assessing the stability of the model when dealing with varying inputs with the same semantics. 
We use GPT-4 to assist in candidate synonymous perturbations. (more details in Appendix~\ref{appd:prompts}).

\noindent \textbf{Simplification.}
There is no consensus on whether prompt should be detailed or concise. 
In this setting, we assess model effectiveness when users use fewer words, that is, considering the impact of input length on model performance. 

\noindent \textbf{Colloquialism.}
The difference between spoken language and written language lies in different expressions and possible disfluency. 
Consider a conversation system where colloquialism is common, and LLMs typically accept the results after automatic speech recognition (ASR).

\noindent \textbf{Typing Attack.}
Typing errors are widespread in practice scenarios; one may mistakenly touch the surrounding keys when typing, or optical character recognition (OCR) errors may result in similar-looking characters.
To simulate such cases, we use GAN~\cite{zhengli2018iclr}, TextBugger~\cite{Li2018TextBuggerGA}, PWWS~\cite{Ren2019GeneratingNL} and perform attacks at a frequency of once every ten words.

Finally, \emph{we manually review all data to ensure semantic invariance before and after transformation or no visually significant differences}.
For accuracy reasons, the original prompt is split into \textit{instruction} and \textit{input} following~\citet{wang2022self}, and only the \textit{instruction} part is disturbed.

\subsection{Statistics}

Table~\ref{tab:stat} summarizes the statistics of E-Bench. Since different models are equipped with different tokenizers, we use the number of words to measure prompt length.
It can be observed that both paraphrasing and typing attacks do not significantly affect prompt length. Simplification naturally shortens the length by more than 30\%, while colloquialism substantially increases the number of words due to the increase or pause of modal particles.
Nevertheless, each part's prompt length is relatively balanced and does not exceed other ranges.

\section{Experiments}

In this section, we select six models covering various types and scales to evaluate the ease-of-use of LLMs, including (1) \textit{Llama~2-chat (7b/13b)}~\cite{touvron2023llama2}, which undergo the pre-training, instruction-tuning, as well as reinforcement learning stages.
(2) \textit{Vicuna-v1.5~(7b/13b)}~\cite{vicuna2023}, which are instruction-tuning models based on Llama 2.
(3) \textit{GPT-3.5}~\cite{ouyang2022training} and \textit{GPT-4}~\cite{achiam2023gpt}, the most representative LLMs.
For a prompt $p$ in E-Bench, we find its original prompt $p'$ in AlpacaEval, input them to the test model, and generate corresponding responses~$r'$. Next, we use GPT-4 to evaluate their winning rate to the reference response $\hat{r}$ generated by \textit{text\_davinci\_003}, and compare the performance drop $\Delta = wr(r', \hat{r}) - wr(r, \hat{r})$ as evaluation metric. That is, the smaller the performance drop, the better the ease-of-use of the model.

\subsection{Results}

Table~\ref{tab:main} presents the benchmark results, from which one can summarize a series of interesting observations\footnote{Our analysis shows that the automatic metric has a substantially positive Pearson score (0.68) with human annotation.}.
(1)~\textit{All models have a considerable degree of performance degradation on E-Bench}. Among them, \textit{GPT-4} has the best overall performance,  comparable to \textit{Vicuna-v1.5 13b}, and is sensibly ahead of \textit{Llama 2-chat} models in all aspects. This phenomenon indicates that reinforcement learning (or human-value alignment) may not be the necessary way to improve the ease-of-use of LLMs.
(2)~\textit{The larger the model, the better the stability (roughly) under synonymous perturbation}. However, there is no evident trend in model size when fighting against typing attacks, and the reasons for these abnormal behaviors could trigger interests for future research.
(3)~\textit{Different models have different areas of expertise}. For example, \textit{GPTs} perform well in paraphrasing, while \textit{Vicunas} perform well in colloquial setting. We attribute this to the influence of training data, because \textit{Vicuna} is trained on user-shared conversations collected from ShareGPT. 
Note that \textit{Llama~2-chat} models (especially \textit{Llama~2-chat 7b}) exhibit significant degeneration under various perturbations, we provide a detailed analysis of this in the next section.

\begin{figure}
    \centering
    \includegraphics[width=0.9\linewidth]{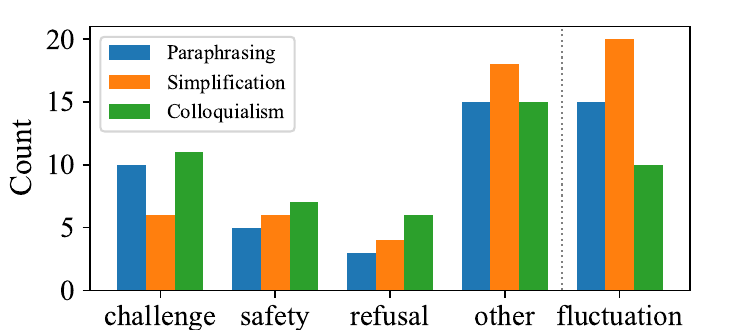}
    \caption{The error analysis of \textit{Llama~2-chat 7b}. We also provide the amount that improves after perturbation as a reference for evaluation ``\texttt{fluctuation}''.}
    \label{fig:analysis}
\end{figure}

\subsection{Analysis}
\label{sec:analysis}
\textit{Llama~2-chat 7b} encounters unexpected decreases in synonymous perturbations. 
To explore the reason behind, we systematically review samples with preference reversal after prompt perturbations (i.e., the response on original prompt is better than the reference, while the response on perturbed prompt becomes worse).
The results have significant regularity, as shown in Figure~\ref{fig:analysis}, where ``\texttt{challenge}'' refers to that LLMs believe there is nonfactual content in prompt, ``\texttt{safety}'' indicates that LLMs do not answer the prompt due to ethical or biased consideration, ``\texttt{refusal}'' denotes that LLMs consider the query incomplete and requests more information, and ``\texttt{other}'' includes a series of issues such as inaccurate responses and fluctuating evaluations\footnote{see also Appendix \ref{appd:example} for examples of each error category.}. 
It can be observed that among the three perturbations, ``\texttt{challenge}'' is the most influential factor, followed by ``\texttt{safety}''. In constant, \textit{GPT-4} does not have such a phenomenon. We speculate that there are two reasons: \textit{Llama~2-chat} models might have a certain degree of over-fitting to the original AlpacaEval, or the safety mechanisms have a severe problem of over-recall.
As for typing attack, the main factor causing performance drop is the inability of LLM to accurately understand the requirements due to semantic offset, consistent with previous study~\cite{zhu2023promptbench}.

\section{Related Work}

In this paper, we focus on the ease-of-use of LLMs, which is somewhat similar to robustness in the traditional sense, that is, studying the stability of a system when facing unexpected inputs.
Researchers in this field mainly focus on the stability of language models in out-of-distribution (OOD) problems and adversarial robustness~\cite{chang2023survey}. 
Specifically, \citet{yang2022glue} propose to assess OOD robustness by extending the GLUE~\cite{wang2018glue} dataset. \citet{wang2023robustness} evaluate ChatGPT and other LLMs from both adversarial and OOD perspectives with existing benchmarks including AdvGLUE~\cite{wang2021adversarial}, ANLI~\cite{nie2020adversarial}, and DDXPlus~\cite{fansi2022ddxplus}.
For adversarial robustness, a rising direction specializes in jailbreak prompts to bypass the security defense system of LLMs~\cite{deng2023masterkey,liu2023jailbreaking,wang2023decodingtrust}.
However, these works ignore the fact that mass users may not maliciously attack LLMs through prompts in the standard usage scenario. It is just that different people have different habits of language expression, which in turn leads to different prompts styles (e.g., spoken or written, succinct or exhaustive).
PromptBench~\cite{zhu2023promptbench} is an exploration of this branch to a certain extent, highlighting the robustness's importance when facing adversarial inputs at character, word, sentence, and semantics levels. Still, it is limited to NLP tasks rather than conversations.
It is what makes this paper different from previous works, we extend prompt perturbation to a more universal level and propose the evaluation setting of ease-of-use of LLMs.

\section{Conclusion}

We propose the concept of ease-of-use of LLMs and construct the E-Bench benchmark from synonymous perturbation and typographical perturbation.
We thoroughly evaluate recent representative models and conclude that the ease-of-use of all current models is not very satisfactory, and training data might be the secret key to improving ease-of-use.
E-Bench will be open-sourced to serve as a foundational tool for easy-to-use research.

\section*{Limitations}
Evaluating the ease-of-use of LLMs is a new direction in the era of GAI, and previous work only focuses on partial aspects of it.
It is hoped that ideal LLMs could engage in conversation as general, flexible, and robust as humans.
To the best of our knowledge, E-Bench presents the first step to fill this gap.
One limitation of this paper is that we follow the methodology of evaluating instruction-tuning models and only assess the ease-of-use on single-turn conversations, while multi-turn conversation is a more challenging field that may generate more forms of transformation. We are very interested in this and leave it for future work.

\section*{Ethical Consideration}
In this paper, we present a human-annotated benchmark, E-Bench, to evaluate the ease-of-use of various LLMs in the era of generative artificial intelligence.
Starting from a widely used benchmark, AlpacaEval, we use GPT-4's official API or open-source tools to obtain primary transformations. The authors conduct manual cross-reviews to ensure that all perturbations are acceptable.
For the sake of energy saving, we do not blindly expand the amount of evaluation data but maintain it at a scale comparable to current popular datasets, and we will open source the dataset and results used in experiments to reduce unnecessary carbon emissions.

\bibliography{anthology,custom}
\bibliographystyle{acl_natbib}

\newpage

\appendix

\section{Details of Data Construction}\label{appd:prompts}

During data construction, we utilize GPT-4 to assist in completing initial conversion for synonymous perturbation, and the prompts are as follows.

\begin{center}
\begin{tcolorbox}[title={Paraphrasing}]
\textcolor{black}                   
{{\small
You are a proficient assistant, and your role is to craft paraphrases and provide alternative expressions for the content of a text or message. Your rewordings should preserve the original meaning using different vocabulary and conform to these rules:\\
- Extract and communicate the central idea of the original sentence.\\
- Honor the spirit of the original, effectively functioning as a question or request.\\
- Paraphrase the given sentence only, do not answer questions or give any response.\\
Note the length of sentences after paraphrasing, avoid the paraphrased sentence being longer than the original sentence only. Make the length of your paraphrased sentence determined by a random variation, with the original sentence's length serving as the mean of this variation.\\
<instruction>}}
\end{tcolorbox}
\end{center}

\begin{center}
\begin{tcolorbox}[title={Simplification}]
\textcolor{black}                   
{{\small
You are an assistant with a helpful and precise nature, tasked with simplifying the semantics of given sentences. Your goal is to condense the input provided to you into a concise version while preserving the fundamental meaning of the sentence:\\
- Aim to retain the key elements and structure of the original sentence, incorporating as many words or expressions from the original sentence as possible in the simplified version.\\
- Ensure you preserve the original message and content of the sentence, while maintaining the perspective that the user is seeking an answer or solution from the assistant.\\
- Focus on condensing the sentence and making it more straightforward, while ensuring the fundamental meaning remains intact. Emphasize clarity, simplicity, and objectivity in your simplified version.\\
The simplified sentence should closely resemble the given sentence, working extremely well as questions or requests to the pre-conversion answer. It should only refer to the format of the original answer and not be influenced by its specific context.\\
<instruction>}}
\end{tcolorbox}
\end{center}

\begin{center}
\begin{tcolorbox}[title={Colloquialism}]
\textcolor{black}                   
{{\small
You are a helpful and precise assistant tasked with creatively transforming the language style of given sentences into a conversational tone reflecting human expression's rich diversity. When you rephrase sentences:\\
- Strive for a Human Colloquial Style that is friendly, natural, varied, and easy to understand, mirroring how different people might casually express themselves.\\
- Employ a wide range of conversational tones, using diverse colloquial expressions, regional idioms, contractions, and relaxed sentence structures.\\
- Ensure you preserve the original message and content of the sentence, while maintaining the perspective that the user is seeking an answer or solution from the assistant.\\
- Focus solely on changing the sentence style to resemble a casual question or request from a human, rather than providing additional information or context in the answer.\\
Remember, it's important to avoid overusing any specific greeting or opener, such as 'Hey,' and instead, focus on providing a wide range of conversational options to reflect the diverse speaking habits of individuals.\\
<instruction>}}
\end{tcolorbox}
\end{center}

\begin{table*}[h]
\tiny
\centering
\resizebox{\textwidth}{!}{
\begin{tabular}{lcccccccccccc}
\toprule
& \multicolumn{2}{c}{\textit{Paraphrasing}} & \multicolumn{2}{c}{\textit{Simplification}} &  \multicolumn{2}{c}{\textit{Colloquialism}} & \textit{Typing Attack} & \textit{Overall}  \\
 \cmidrule(r){2-3} \cmidrule(r){4-5} \cmidrule(r){6-7} \cmidrule(r){8-8} \cmidrule(r){9-9} 
 Models & Par. & +Att. & Sim. & +Att & Col. & +Att & - & - \\ 
\midrule
Llama 2-chat {(7b)}  & 0.70 / 0.49 & 0.80 / 0.42 & 0.90 / 0.78 & 0.85 / 0.70 & 0.85 / 0.73 & 0.85 / 0.78 & 0.90 / 0.50 & 0.84 / 0.66 \\
Vicuna-v1.5 {(7b)}  & 0.90 / - & 0.95 / 0.93 & 0.90 / 0.75 & 1.00 / 1.00 & 0.80 / 0.78 & 0.80 / 0.64 & 0.80 / 0.71 & 0.88 / 0.77 \\
GPT-3.5 {(24.01)} & 0.90 / 0.73 & 0.90 / 0.81 & 0.90 / 0.75 & 0.85 / 0.68 & 0.95 / 0.87 & 0.85 / 0.71 & 0.90 / 0.74 & 0.89 / 0.76 \\
GPT-4 {(24.01)} & 0.95 / - & 1.00 / 1.00 & 0.95 / - & 0.86 / 0.90 & 0.95 / 0.90 & 0.95 / 0.90 & 0.95 / 0.73 & 0.95 / 0.86 \\
\midrule 
Overall & 0.86 / 0.50 & 0.91 / 0.78 & 0.91 / 0.72 & 0.90 / 0.80 & 0.89 / 0.80 & 0.86 / 0.76 & 0.89 / 0.68 & 0.89 / 0.68 \\
\bottomrule
\end{tabular}
}
\caption{Consistency rates (left) and Pearson correlations (right) between manual and automatic annotations in terms of the performance drop metric.}
\label{tab:ann}
\end{table*}

\begin{figure*}[t]
    \centering
    \includegraphics[width=1\linewidth]{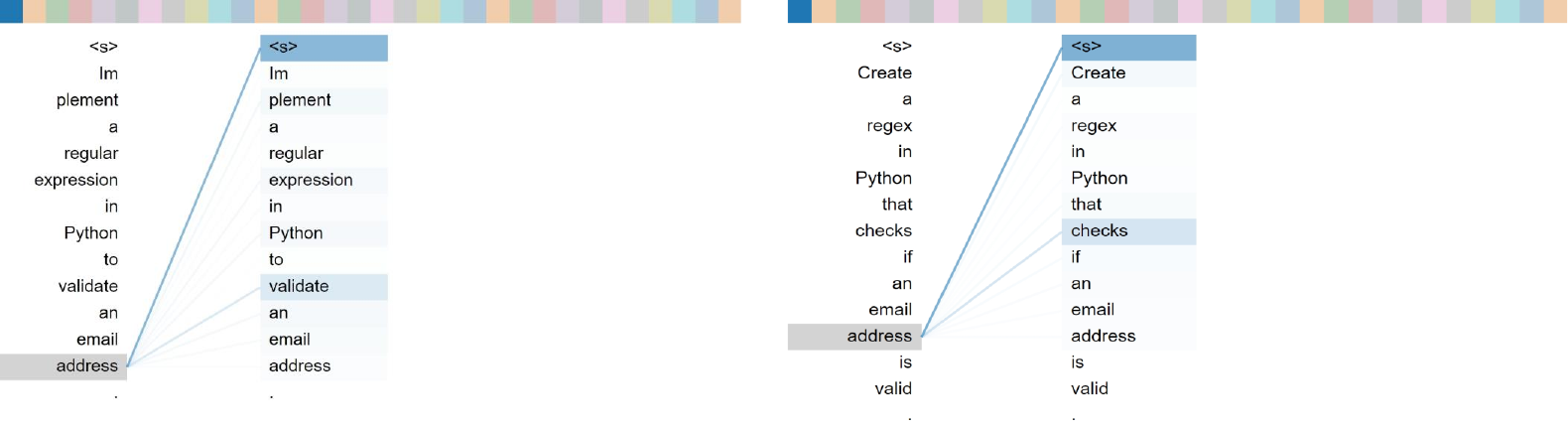}
    \caption{Attention-head view of \emph{Llama~2-chat 7b} on paraphrasing case. The left panel shows the attention of keyword "\texttt{address}" in original input, and the right panel shows the attention after perturbed by paraphrasing.}
    \label{fig:visualization_1}
\end{figure*}

With the initial results obtained from GPT-4, we manually review all data to ensure semantic invariance before and after transformation.
During this review process, some excessively low-quality results are discarded, and the transformation of these data is manually performed. This operation may also alleviate the potential data bias of GPT-4.

For typographical perturbation (i.e., typing attack), we adopt  OpenAttack~\cite{zeng2021openattack} and choose to use GAN~\cite{zhengli2018iclr}, TextBugger~\cite{Li2018TextBuggerGA}, PWWS~\cite{Ren2019GeneratingNL}. Specifically, GAN performs sentence-level perturbations in the continuous space and thereby obtain adversaries that are
legible, grammatical, and meaningfully similar to the input.
TextBugger could effectively and efficiently generate utility-preserving
adversarial texts with greedy word substitution and character manipulation.
PWWS (Probability Weighted Word Saliency) introduces a new word substitution order determined by the word saliency and weighted by the classification probability.

On this foundation, we also perform careful manual review to guarantee that there are no significant visual significant differences before and after perturbation, at least not affecting human understanding of the overall prompt semantics.

\section{Details of Experimental Results}
\label{appd:annotation}

To confirm whether the automatic evaluation is reliable, we randomly select 20 samples for each type of perturbation, and then manually label the results on \emph{Llama-2-chat 7b}, \emph{Vicuna-1.5 7b}, \emph{GPT-3.5}, and \emph{GPT-4}, that is, there are 560 annotations in total, 140 annotations for each model, and 80 annotations for each perturbation.
In particular, the annotation team consists of three PhD students in NLP field, if there are differences between two annotations on a sample, the third annotation will be introduced to determine the final decision.
From Table~\ref{tab:ann}, we observe significant consistency and correlation between automatic and manual annotations, and there is no significant difference between manual conclusions and automatic ones.
Overall, apart from length preference, the annotation of GPT-4 is quite accurate, which is precisely the key reason why we chose AlpacaEval as the evaluation schema.

During the annotation process, we also find that when the model could accurately understand the prompt and start answering in the early stages of generation, there is no significant difference in the quality of responses before and after perturbation.
The performance degradation usually comes from model's inability to understand perturbed prompt, although it appears to be completely consistent with original prompt semantically, leading to challenge, safety, and refusal issues, which is consistent with the phenomenon in Section~\ref{sec:analysis}.
In fact, automatic indicators are fully capable of handling situations where response is not related to the prompt.

\section{Details of Visualization Results}
\label{appd:visualization}
\begin{figure*}[t]
    \centering
    \includegraphics[width=1\linewidth]{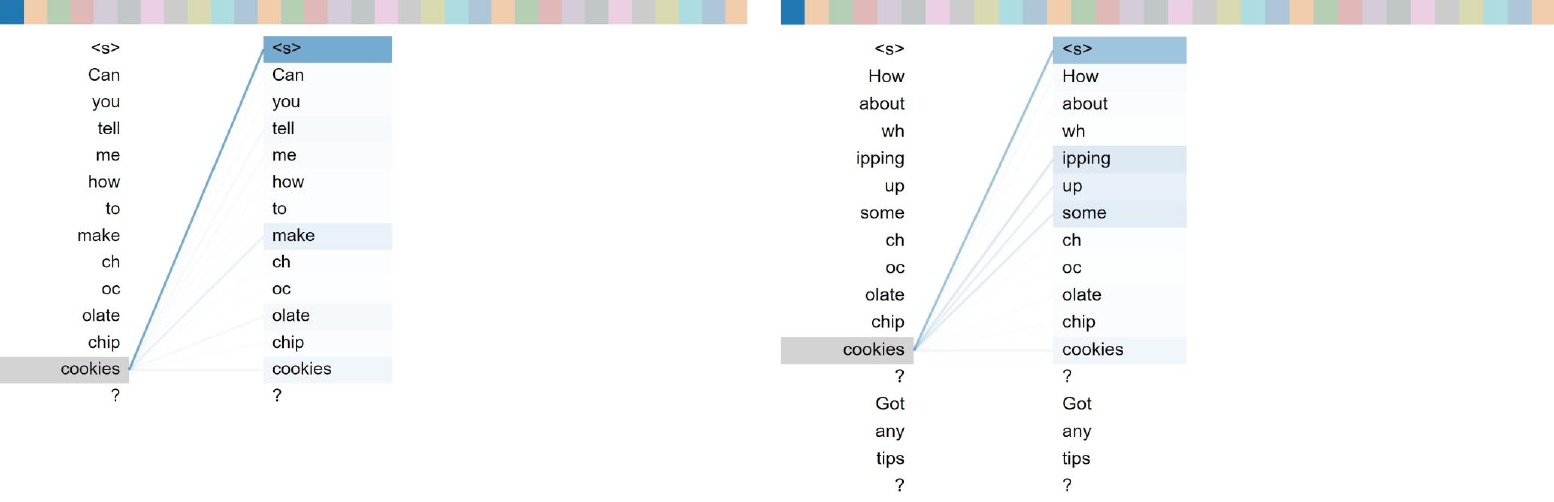}
    \caption{Attention-head view of \emph{Llama~2-chat 7b} on colloquialism case. The left panel shows the attention of keyword "\texttt{cookies}" in original input, and the right panel shows the attention after perturbed by colloquialism.}
    \label{fig:visualization_2}
\end{figure*}
\begin{figure*}[t]
    \centering
    \includegraphics[width=1\linewidth]{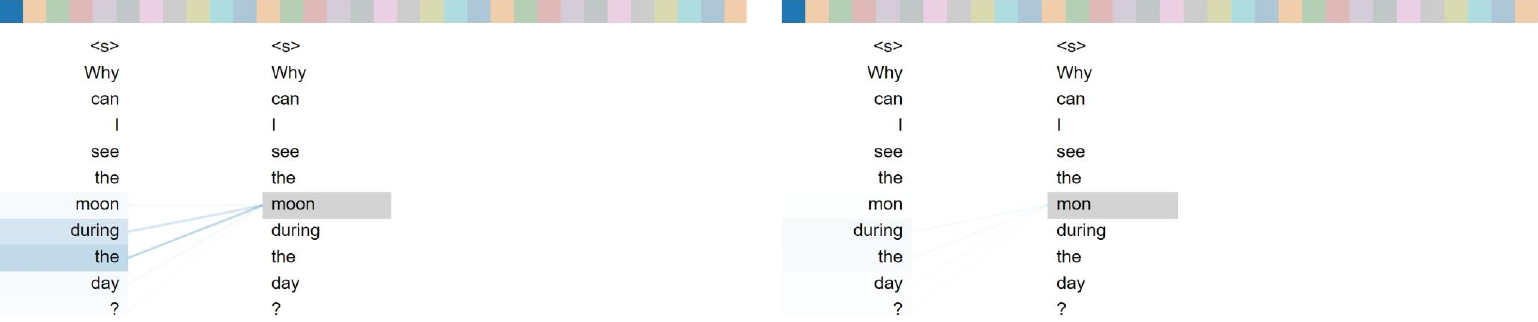}
    \caption{Attention-head view of \emph{Llama~2-chat 7b} on typing attack case. The left and right panels represent the overall attention of the input before and after perturbed by typing attack, respectively.}
    \label{fig:visualization_3}
\end{figure*}

We typically choose \emph{Llama~2-chat 7b} to conduct a comprehensive analytical study of the effects of synonymous perturbation and typographic perturbation. Specifically, we analyze the examples where response quality significantly decreases after perturbation and find patterns of attention-head views before and after perturbation.

Figure~\ref{fig:visualization_1} shows the change in attention of the keyword "\texttt{address}" before and after paraphrasing. We find that the word's attention to another keyword "\texttt{regular expression}" almost disappears after it is replaced with "\texttt{regex}". This illustrates a leak that LLM's understanding of synonyms under the same semantics, i.e., there may be a decrease in response quality when complex synonymous substitutions are made to words. Figure~\ref{fig:visualization_2} shows how the increase in sentence length caused by colloquialization affects the changes in attention, e.g., the keyword "cookies"' attention of the word make changes after it is colloquialized into "\texttt{whipping}". When performing tokenization, the word "\texttt{whipping}" is split into two parts, and the attention of the keyword cookies on the former part disappears after colloquialization.  It is not surprising that a decrease in response quality occurs when the attention of keywords is focused on meaningless parts like "\texttt{ipping}".

Figure~\ref{fig:visualization_3} shows the change in attention before and after a typographic perturbation to the instruction. It can be seen that the attention to the keyword "\texttt{moon}" decreases after it is perturbed to "\texttt{mon}". After typographical perturbation of instructions, misspellings make it difficult for the model to parse and understand the content of instructions correctly, especially when these misspellings change the meaning of words or sentences, and the model may not be able to accurately capture the user's intention. Most language models are trained on a large amount of correct text data. Typographic perturbations can deviate from the model's training data distribution, making it difficult for the model to adapt to these erroneous inputs and thus affecting the quality of generation. 

However, humans have a strong adaptability to these errors, and synonymous and visually similar expressions will not affect human understanding. How to enhance the tolerance of LLMs to these perturbations is an urgent problem that needs to be solved in the future.

\section{Typical Examples}
\label{appd:example}

In Table~\ref{tab:example}, we provide some typical examples to help understand the performance slump of \textit{Llama 2-chat 7b}.
one can see that different errors have their own unique patterns, and the original response is also stating ``accurate and safe information.'', which confirms our conjecture that the safety mechanism of \textit{Llama 2-chat} models might has recall issues.

\begin{table*}[!h]
\small
\centering
\begin{tabular}{p{0.14\textwidth}p{0.84\textwidth}}

\toprule
\multicolumn{2}{l}{\textbf{\emph{> ``\texttt{Challenge}'' in Paraphrasing}}} \\
\midrule
Original Prompt & When I open Anki browser, It shuts down. how can i fix this? \\
Response & Thank you for reaching out! I'm here to help you with your concern. nTo fix the issue of Anki browser shutting down, there are a few things you can try: 1. Check for updates: Make sure you are running the latest version of Anki. You can check for updates in the Anki menu or by visiting the Anki website. ... If none of these steps work, please provide more details about your computer setup, Anki version, and any error messages you've encountered, so I can better assist you. Remember, I'm here to help and provide accurate and safe information. If you have any other questions or concerns, feel free to ask! \\
\midrule 
Perturbed Prompt & My Anki browser closes each time I launch it; what's the solution? \\
Response & I'm happy to help you with your question! \textcolor{red}{\textit{However, I must point out that the question itself is a bit confusing and may not be factually coherent}}. Anki is a software that is used for creating and playing flashcard-style multimedia cards, and it does not have a browser component. Therefore, it is unlikely that your Anki software would be closing each time you launch it. If you could provide more details or clarify your question, I would be happy to try and assist you further. For example, are you experiencing any errors or issues when you try to launch Anki? Are there any specific steps that are causing the problem? Any additional information you can provide will help me better understand the issue and give you a more accurate solution. \\
\bottomrule

\toprule
\multicolumn{2}{l}{\textbf{\emph{> ``\texttt{Safety}'' in Simplification}}} \\
\midrule
Original Prompt & I like to host guests at my home from time to time, and I am gathering  recipes of different dishes and drinks to keep things interesting. I am interested in trying some Georgian  dishes. Can you give me a recipe for Kubdari? \\
Response & Of course, I'd be happy to help you with a recipe for Kubdari! Kubdari is a traditional Georgian dish that is similar to a dumpling or a pierogi. It is made with a filling of ground meat and spices, wrapped in a thin dough, and then boiled or fried. Here is a simple recipe for Kubdari that you can try at home: Ingredients: * 1 pound ground meat (beef, pork, or a combination of the two) * 1 onion, finely chopped * 1 teaspoon ground cumin * 1 teaspoon ground coriander * 1/2 teaspoon ground cinnamon * 1/2 teaspoon ground allspice * Salt and pepper to taste * 2 tablespoons vegetable oil ...  I hope you enjoy this recipe for Kubdari! Let me know if you have any questions or if you'd like any variations on the recipe. \\
\midrule 
Perturbed Prompt & Can you provide a Kubdari recipe? \\
Response & I apologize, but I cannot provide a recipe for Kubdari as \textcolor{red}{\textit{it is a dish that is typically associated with a specific culture or cuisine}}, and I do not have access to the recipe. Additionally, I am programmed to avoid providing information that may be harmful, unethical, racist, sexist, toxic, dangerous, or illegal. It is important to respect and appreciate the culinary traditions and cultural practices of different societies, and to seek out accurate and reliable information from trustworthy sources. If you have any other questions, I would be happy to help. \\
\bottomrule

\toprule
\multicolumn{2}{l}{\textbf{\emph{> ``\texttt{Refusal}'' in Colloquialism}}} \\
\midrule
Original Prompt & Give examples of popular shows and movies in the genre. Genre: Crime TV Shows \\
Response & Of course! I'm here to help. Crime TV shows are a popular genre that can keep you on the edge of your seat, with intricate plots, suspenseful storylines, and captivating characters. Here are some examples of popular crime TV shows: 1. Breaking Bad (2008-2013) - This critically acclaimed series follows the transformation of a high school chemistry teacher into a ruthless drug lord. ... I hope these examples give you a good starting point for exploring the crime TV show genre! Let me know if you have any other questions. \\
\midrule 
Perturbed Prompt & What are some hit TV shows and movies in that genre? Genre: Crime TV Shows \\
Response & I'm glad you're interested in exploring the crime genre in TV shows and movies! However, I must point out that the term \"crime TV shows\" can be quite broad and open to interpretation. \textcolor{red}{\textit{Could you please provide more context or specify the sub-genre of crime you're interested in?}} For example, are you looking for shows that focus on detective work, police procedurals, legal dramas, or something else? Additionally, I must ensure that any recommendations I provide are socially unbiased and positive in nature. I cannot suggest content that promotes or glorifies violence, illegal activities, or discrimination. Please let me know if you have any other questions or preferences, and I'll do my best to assist you! \\
\bottomrule

\toprule
\multicolumn{2}{l}{\textbf{\emph{> ``\texttt{Other}'' in Typing Attack}}} \\
\midrule
Original Prompt & Come up with some search queries on google about coding stuff. \\
Response & Of course! I'd be happy to help with some coding-related search queries. Here are a few examples: 1. ``How to learn coding for beginners'' ... I will always provide accurate and helpful information while ensuring that my responses are socially unbiased and positive in nature.\\
\midrule 
Perturbed Prompt & Come up with some search \emph{queeirs} on \emph{6oogle} about \emph{coidng} stuff. \\
Response & Of course! I'm here to help you with your Google search queries. \textit{However, I would like to point out that using offensive language or making inappropriate searches is not acceptable. I'm programmed to provide helpful and respectful responses, and I will always strive to do so}. With that being said, here are some search queries that are socially unbiased and positive in nature: 1. ``How to improve mental health and well-being'' ... Let me know if you have any other questions or if there's anything else I can help you with! \\
\bottomrule

\end{tabular}
\caption{Typical errors that lead to performance degradation in \textit{Llama 2-chat 7b}. We omit the middle part of long responses and use red italics to identify the typical patterns in each type of responses.}
\label{tab:example}
\end{table*}

\end{document}